\pdfoutput=1
\documentclass[runningheads]{llncs}

\usepackage{url}

\usepackage{amsmath}
\usepackage{amssymb}

\usepackage{cellspace}

\usepackage{pgfplots}
\usepackage{caption}

\usepackage{mathtools, nccmath}

\usepackage{tikz}
\usetikzlibrary{shapes}

\usepackage{algorithm}
\usepackage{algpseudocode}

\usepackage{multirow}
\usepackage{subcaption}
\usepackage{colortbl}
\usepackage{tabularray}
\usepackage{booktabs}
\usepackage[table]{xcolor}
\definecolor{cvprblue}{rgb}{0.21,0.49,0.74}
\usepackage[pagebackref,breaklinks,colorlinks,allcolors=cvprblue]{hyperref}

\tikzset{
circ/.style={draw,circle,minimum height=3em},
}

\begin{document}
\setcounter{MaxMatrixCols}{50}

\title{Diffusion-Based Synthetic Brightfield Microscopy Images for Enhanced Single Cell Detection}
\titlerunning{Diffusion-Based Synth. Brightf. Microscopy Images for Enh. Single Cell Det.}

\author{Mario de Jesus da Graca\inst{1} \and Jörg Dahlkemper\inst{2} \and Peer Stelldinger\inst{2}\\
\institute{ Synentec GmbH, Elmshorn, Germany \and Faculty of Computer Science and Digital Society, HAW Hamburg, Germany} 
\email{m.dagraca@synentec.com}\\
\email{joerg.dahlkemper@haw-hamburg.de}\\
\email{peer.stelldinger@haw-hamburg.de}
}
\authorrunning{M. da Graca, J. Dahlkemper and P. Stelldinger}
\maketitle              % typeset the header of the contribution

\begin{abstract}
   Accurate single cell detection in brightfield microscopy is crucial for biological research, yet data scarcity and annotation bottlenecks limit the progress of deep learning methods.  We investigate the use of unconditional models to generate synthetic brightfield microscopy images and evaluate their impact on object detection performance. A U-Net based diffusion model was trained and used to create datasets with varying ratios of synthetic and real images. Experiments with YOLOv8, YOLOv9 and RT-DETR reveal that training with synthetic data can achieve improved detection accuracies (at minimal costs). A human expert survey demonstrates the high realism of generated images, with experts not capable to distinguish them from real microscopy images (accuracy 50\%). Our findings suggest that diffusion-based synthetic data generation is a promising avenue for augmenting real datasets in microscopy image analysis, reducing the reliance on extensive manual annotation and potentially improving the robustness of cell detection models.
\end{abstract}

\section{Introduction}
\label{sec:introduction}
Single cell detection in microscopy images is a fundamental task in biological and medical applications, enabling quantitative analysis of cellular processes and disease mechanisms~\cite{meijering_cell_2012}.
Accurate identification and localization of individual cells are crucial for understanding cellular heterogeneity, dynamics, and responses to stimuli, driving advancements in fields ranging from drug discovery to diagnostics~\cite{f_mualla_automatic_2016,e_d_ferreira_classification_2024}.
Brightfield microscopy, a label-free technique, is widely used due to its non-invasiveness and simplicity, allowing for long-term observation of live cells without phototoxicity or alteration of cellular states~\cite{jyrki_selinummi_bright_2009,huixia_ren_cellbow_2020}.
However, brightfield microscopy presents significant challenges for automated analysis due to inherent limitations such as low contrast between cells and background, high variability in cell appearance influenced by factors like cell type and imaging conditions, and the presence of artifacts like uneven illumination and well edge distortions~\cite{jyrki_selinummi_bright_2009,f_mualla_automatic_2016}.
These challenges often necessitate labor-intensive manual annotation, hindering high-throughput analysis and objective analysis.

Deep learning approaches, particularly CNN-based object detectors, have shown great promise for automatic cell detection~\cite{erick_moen_deep_2019,thorsten_falk_u-net_2019}.
However, the performance of these models is heavily dependent on large, annotated datasets, which are expensive and time-consuming to acquire in microscopy~\cite{ronneberger_u-net_2015}.
Moreover, ethical considerations arising from the use of biological samples derived from human subjects or animal models require complex approval processes, like obtaining informed consent, confidentiality of sensitive information and compliance to regulations on the use and storage of biological material~\cite{anne_cambonthomsen_series_2007,i_galende_ethical_2023,gregory_pappas_exploring_2005}.
These requirements further complicate the acquisition of sufficient real-world data, especially for \textit{in vitro} studies involving diverse cell lines and experimental conditions.

Synthetic data generation offers a potential solution to alleviate data scarcity.
Recent advancements in diffusion models have demonstrated their ability to generate high-fidelity images across various domains~\cite{ho_denoising_2020,song_denoising_2020}.
In microscopy, synthetic data could augment real datasets, improve model generalization, and reduce annotation efforts~\cite{rajaram_simucell_2012,trampert_deep_2021,lehmussola_synthetic_2008}.
Diffusion models, unlike GANs, offer improved training stability and the ability to generate diverse samples, making them particularly attractive for capturing the inherent variability in cell morphology and imaging artifacts often encountered in brightfield microscopy.

This paper explores the use of unconditional diffusion models for generating synthetic brightfield microscopy images of CHO cells and investigates their impact on single cell detection accuracy using state-of-the-art object detectors.
The diffusion model was trained on a carefully curated dataset of 10,000 images (512x512 pixels) of stably transfected CHO-K1 and CHO DG44 cell lines, acquired using a high-throughput microscope.
We address the central research question: \textbf{Can synthetic brightfield microscopy images generated by diffusion models be used to enhance the performance of object detection models for single cell detection?}
We further investigate: (1) the perceptual realism of generated images through expert evaluation, and (2) the influence of synthetic data proportion in training datasets.

Our contributions include: (i) application of unconditional diffusion models for realistic brightfield microscopy image synthesis;
(ii) a comprehensive evaluation of synthetic data augmentation for single cell detection using state-of-the-art object detectors (YOLOv9, YOLOv9, RT-DETR);
(iii) expert validation of generated image realism;
and (iv) insights into the potential and limitations of diffusion-based synthetic data for microscopy image analysis.

\section{Related Work}
\label{sec:related-work}
Synthetic microscopy image generation has become an increasingly important area of research, driven by the need for large, annotated datasets for training and validating image analysis algorithms.
Generating synthetic data offers a promising avenue to overcome the limitations of acquiring and annotating real microscopy images, which are often expensive, time-consuming, and require expert knowledge.
Early approaches to synthetic microscopy image generation relied on rule-based models and physical simulations~\cite{david_svoboda_generation_2012,david_svoboda_towards_2013,rajaram_simucell_2012}.
These methods, while offering control over image parameters and ground truth generation, were often limited in their ability to capture the full complexity and biological diversity observed in real-world microscopy data.

Early works in synthetic microscopy image generation focused on creating controlled datasets for algorithm development and validation.
Svoboda et al. pioneered in the generation of fully 3D synthetic brightfield microscopy time-lapse sequences, modeling cell shapes, structures, and motion~\cite{david_svoboda_generation_2012}.
This work was further extended to simulate realistic cell population distributions in 3D by controlling cell count and clustering probability~\cite{david_svoboda_towards_2013}.
SimuCell, developed by Rajaram et al.~\cite{rajaram_simucell_2012}, is another example of a simulation-based approach, aiming to mimic brightfield microscopy images for cell segmentation algorithm development.
These methods, based on predefined rules and physical models, provided valuable initial datasets for testing algorithms, particularly for segmentation and tracking.
However, their inherent limitations in replicating the intricate textures, subtle variations, and artifacts present in real biological images restricted their utility for training more sophisticated, data-driven models, especially deep learning approaches.

The advent of deep learning, particularly Generative Adversarial Networks (GANs), marked a significant shift in synthetic image generation.
GANs offered the potential to learn complex data distributions directly from real images, leading to more realistic and diverse synthetic ouputs.
In the context of microscopy, GANs have been extensively explored for various tasks.
Cross-modality translation, aiming to predict fluorescence microscopy images from brightfield inputs, has been a prominent application.
Christiansen et al.\ introduced ``In Silico Labeling'' (ISL) using deep learning to predict fluorescence labels from transmitted light images~\cite{christiansen_silico_2018}.
Building upon this, Lee et al.\ developed DeepHCS and DeepHCS++ to transform brightfield images into multiple fluorescence channels relevant for high-contrast screening, leveraging adversarial losses for realism~\cite{gyuhyun_lee_deephcs_2018,gyuhyun_lee_deephcs_2021}.
GANs have also been applied to enhance image quality achieving super-resolution under a large field-of-view~\cite{zhang_high-throughput_2019} and generate isolate cell images approximating the shape and texture distributions of cell components~\cite{marin_scalbert_generic_2019}.
Despite these successes, GANs are known to suffer from training instability and mode collapse, limiting the diversity and reliability of generated samples, particularly when training data is scarce, complex or imbalanced, as often encountered in microscopy~\cite{juan_c_caicedo_evaluation_2019,ricard_durall_combating_2020}.

Diffusion models, inspired by non-equilibrium thermodynamics and rooted in denoising score matching, offer improved training stability, sample quality and diversity~\cite{jascha_sohl-dickstein_deep_2015,ho_denoising_2020,song_denoising_2020}.
These models have demonstrated state-of-the-art performance in various image generation tasks and are increasingly being explored in microscopy:
Cross-Zamirski et al.\ introduced a class-guided diffusion model to generate cell painting images from brightfield data, along with class labels, demonstrating the potential for controlled image synthesis~\cite{cross-zamirski_class-guided_2023}.
Della Maggiora et al.\ presented a conditional variational diffusion model for super-resolution microscopy and quantitative phase imaging, focusing on improving the noise schedule learning during training~\cite{gabriel_della_maggiora_conditional_2023}.
In electron microscopy, Lu et al.\ developed EMDiffuse, a suite of algorithms based on diffusion models for image enhancement, reconstruction, and generation, showcasing the versatility of these models across different microscopy modalities~\cite{lu_diffusion-based_2024}.
Li et al.\ proposed a physics-informed diffusion model (PI-DDPM) that incorporates the physical model of microscopy image formation into the loss function, leading to improved reconstructions with reduced artifacts compared to regular DDPMs when trained on synthetic microscopy data~\cite{li_microscopy_2023}.
These works highlight the growing interest in diffusion models for various microscopy image synthesis and enhancement tasks, demonstrating their capability to generate high-quality and diverse samples.

While diffusion models are increasingly used in microscopy image generation, research on their direct application for enhancing \textbf{object detection}, specifically in the challenging domain of brightfield microscopy, remains significantly limited.
Existing studies primarily focus on image generation quality evaluation and downstream tasks like segmentation or image-based profiling.
The critical question of whether and how diffusion-based synthetic brightfield microscopy images can effectively improve the performance of object detection models for single cell detection has not been systematically addressed.
This work is directly pertinent to the aforementioned issue by undertaking a comprehensive evaluation of the impact on diffusion-based synthetic brightfield microscopy image data on single cell detection performance.
By focusing on object detection, this research contributes a novel perspective to the field, exploring the practical utility of synthetic data for a fundamental task in quantitative microscopy.

\section{Methodology}
\label{sec:methodology}
Our methodology is structured into three distinct yet interconnected stages.
These stages are: (1) training and unconditional diffusion model and generating synthetic brightfield microscopy images, (2) training and evaluating state-of-the-art object detection models using two sets of datasets with varying proportions of synthetic images replacing and adding real images, and
(3) conducting an expert survey to assess the perceptual realism of generated images.
This multi-faceted approach allows for a comprehensive investigation into the efficacy of synthetic data augmentation in the context of cell detection.

\subsection{Image Generation with Diffusion Models}
\label{subsec:image-generation-with-diffusion-models}
The first stage of our methodology focuses on generating realistic synthetic brightfield microscopy images using an unconditional diffusion model.
This stage involves dataset preparation, diffusion model training, model selection, and the final image generation process.

The diffusion model was trained using a dataset of 10,000 patches ($512 \times 512$) extracted from real brightfield microscopy images.
These images were acquired using a CELLAVISTA 3.1 RS HE high-throughput microscope (Synentec GmbH) with a $10\times$ objective lense, resulting in 8-bit grayscale images.
The source material consisted of stably transfected CHO cells (CHO-K1 and CHO DG44 variants) expressing eGFP, imaged in 96-well plates at three different densities (300, 150, and 1 cell per well).

Prior to training, raw subwell images ($3056 \times 3056$ pixels) were patchified into $512 \times 512$ pixel non-overlapping patches to increase the dataset size and focus the model on local cell features.
Patches containing visible well edges, identified as dark arch artifacts due to light refraction at the well boundaries, were manually excluded to ensure the training data primarily comprised relevant cellular structures.
The final dataset of 10,000 patches was randomly sampled from the filtered set of approximately 24,000 patches to manage computation resources within the scope of this research.
For efficient data loading and processing during training, the dataset was converted into the Hugging Face Datasets format, stored as parquet files, which facilitates faster data access and management~\cite{vohra_apache_2016}.

We employed unconditional diffusion models based on U-Net architectures and attention mechanisms as the backbone for image generation.
The U-Net architecture was chosen for its proven effectiveness in image generation and segmentation tasks, particularly in biomedical imaging, due to its ability to capture both local and global context through its encode-decoder structure and skip connections~\cite{ronneberger_u-net_2015}.
The specific U-Net architecture implemented consists of downsampling and upsampling blocks with residual connections to facilitate gradient flow and improve training stability.
Attention mechanisms were optionally incorporated into both the downsampling and upsampling paths in some model variants to investigate their impact on image generation quality.
The number of up and down blocks, channel configurations and the placement of attention blocks were varied across different model configurations to evaluate architectural choices.

The diffusion model was trained using the DDIM scheduler~\cite{song_denoising_2020} for efficient sampling during training and to explore faster inference capabilities.
The model was trained for 350 epochs using the AdamW optimizer with a learning rate of $10^{-4}$.
Moving Average (EMA) with a decay of 0.9999 was applied to the model weights to stabilize training and potentially improve generalization.
A batch size of 4 was used, constrained by GPU memory limitations.
Training was performed using Distributed Data Parallel (DDP) across two NVIDIA A100-SXM4-40GB GPUs with mixed precision (float16) enabled using PyTorch Lightning~\cite{falcon_pytorchlightningpytorch-lightning_2020} to accelerate training.

Model selection was based on a combination of visual inspection of generated samples and quantitative evaluation using the Fréchet Inception Distance (FID) scores.
FID scores were calculated against a held-out set of real images to objectively measure the similarity between the distributions of real and generated images and computed every 10 epochs to track training progress and identify optimal model states.

For the final synthetic image generation, the Euler Ancestral scheduler was chosen for its balance of sample quality and generation speed, determined through empirical testing of various schedulers.
Trailing timestep spacing and epsilon prediction type were used based on hyperparameter optimization experiments that indicated these settings produced visually superior results for our dataset.
To introduce variability and potentially enhance the robustness of the downstream object detectors, the number of inference steps during the final generation process was randomized between 35 and 40 steps for each batch of 16 images.
A total of 10,000 synthetic images were generated using this optimized configuration to create datasets for object detection model training.

\subsection{Cell Detection Model Training and Evaluation}
\label{subsec:cell-detection-model-training-and-evaluation}
The second stage of our methodology involved training and evaluating state-of-the-art object detection models using datasets incorporating varying proportions of synthetic images.
This stage aimed to quantify the impact of synthetic data augmentation on cell detection performance.

We created six datasets containing synthetic images and a baseline dataset consisting of 5,000 real images (\textit{scc\_real}).
The baseline dataset served as a control to evaluate the performance of detection models trained solely on real data.
To investigate the effect of replacing real images with synthetic ones, three datasets (\textit{scc\_10}, \textit{scc\_30}, \textit{scc\_50}) were created where 10\%, 30\%, and 50\% of the real images in the training set where replaced with synthetic images, respectively.
\textit{scc\_add\_10}, \textit{scc\_add\_30}, \textit{scc\_add\_50} were created with the same proportions of synthetic images added to the training set, keeping the original real images intact and increasing the total dataset size.

The remaining images from the real dataset (beyond the 5,000 training images) were allocated to fixed validation (2,527 images) and test (16,758 images) sets, ensuring consistent evaluation across all training configurations.
These large validation and test sets kept constant across all experiments to provide robust and statistically meaningful performance comparisons.

Real images were labeled using a semi-automated approach that combined fluorescence channel information with manual verification.
First, a classical image processing algorithm was implemented to automatically detect cells in the fluorescence channel (eGFP signal).
This algorithm involved binarizing the fluorescence image, finding contours corresponding to fluorescent signals, and generating bounding boxes around these contours.
The automatically generated bounding boxes were then manually reviewed and refined by a human expert using the Roboflow annotation tool~\cite{dwyer_roboflow_nodate} to ensure accuracy and correct any errors from the automated process.
This hybrid approach significantly reduced the manual labeling effort while maintaining high-quality annotations.

Synthetic images were labeled using a model-assisted labeling approach.
A pretrained YOLOv8m model was fine-tuned on a subset of real labeled images.
This fine-tuned YOLOv8m model was integrated into the Roboflow annotation tool's ``Model-Assisted Labeling'' feature.
These model-predicted bounding boxes were also manually reviewed and refined by a human expert to correct any inaccuracies, missed detections or false positives, to ensure consistency between real and synthetic image labelling.
This approach leveraged the pre-trained object detection model to expedite the labeling of synthetic data, while manual refinement ensured the final labels were of high quality.

We fine-tuned three state-of-the-art object detection model families: YOLOv8 (sizes s, m, x)~\cite{jocher_ultralytics_2023}, YOLOv9 (sizes c, e)~\cite{Wang2024YOLOv9LW}, and RT-DETR (sizes l, x)~\cite{Lv2023DETRsBY}.
These models were chosen for their state-of-the-art performance, diverse architectures (CNN-based and Transformer-based), and availability within the Ultralytics framework~\cite{jocher_ultralytics_2023}, which facilitated consistent training and evaluation.
All models were pre-trained on the COCO dataset~\cite{tsung-yi_lin_microsoft_2014} and fine-tuned on our cell detection datasets to leverage transfer learning, accelerate training and reduce the amount of data required for training.

Models were trained for 200 epochs using default Ultralytics augmentation settings, which include mosaic augmentation, mixup augmentation, and color space augmentations (HSV) to enhance model robustness and generalization.
An early stopping mechanism with a patience of 35 epochs was employed to prevent overfitting and optimize training time.
Training was performed on a single NVIDIA GeForce RTX 3090 GPU with 24GB of memory.

Model performance was evaluated on the held-out test set of 16,758 real images.
Performance metrics included mean Average Precision (mAP) at IoU thresholds of 0.5 (mAP@50), 0.75 (mAP@75), and averages across 0.5 to 0.95 (mAP@50:95).
mAP\@50 was chosen as the primary metric for performance comparison, as it provides a robust measure of detection accuracy while being less sensitive to precise boundary delineation, which is less critical in our single cell detection task focused on identifying and localizing cells rather than detailed morphological analysis.
However, mAP\@75 and mAP\@50:95 were also reported to provide a more comprehensive performance profile across different IoU thresholds.

\subsection{Expert Survey for Image Realism}
\label{subsec:expert-survey-for-image-realism}
To subjectively assess the perceptual realism of the generated synthetic images, we conducted a survey involving 11 microscopy experts and biologists from Synentec GmbH.
These experts possess extensive experience in microscopy image analysis and cell biology, making them ideal judges of image realism in this domain.

Participants were presented with 30 randomly ordered images, consisting of 20 synthetic images generated by our diffusion model and 10 real brightfield microscopy images from our dataset, ensuring a balanced representation of both image types without revealing the true ratio to participants.
For each image, participants were asked to clarify it as either ``real'' or ``synthetic'' and to rate their confidence in their classification on a scale from 1 (low confidence) to 5 (high confidence).
Furthermore, if participants classified an image as ``generated'', they were prompted to provide a brief textual explanation detailing the visual cues or artifacts that led to their decision.

We analyzed the classification accuracy of the experts, calculating overall accuracy as well as accuracy separately for real and synthetic images.
We also collected and analyzed the textual explanations provided by participants for images they classified as generated.
These textual explanations were analyzed qualitatively to identify recurring themes and visual features that experts associated with synthetic images.

This comprehensive methodology, combining quantitative evaluations of object detection performance with subjective assessments of image realism by expert human observers, provides a robust framework for investigating the impact of synthetic data augmentation in brightfield microscopy cell detection.

\section{Experiments and Results}
\label{sec:experiments-and-results}
After thorough evaluation of the various unconditional diffusion model architectures, noise schedulers, and hyperparameters,
a final configuration was selected for the image generation task.
The chosen model is the \textit{scc\_small} architecture, which demonstrated a good balance between performance and computational efficiency.
It consists of 5 block out channels (128, 128, 256, 256, 512) and 5 block in channels (128, 128, 256, 256, 512), with no attention layers and a total of 70.1M trainable parameters.
It was trained for 350 epochs in two days and 7 hours, with the above mentioned hyperparameters and noise scheduler settings.
This model, despite its relatively smaller size, achieved competitive results in terms of image quality and fidelity.

\subsection{Expert Survey: Indistinguishability of Synthetic Images}
\label{subsec:expert-survey:-indistinguishability-of-synthetic-images}
\begin{figure}
    \centering
    \includegraphics[width=0.75\linewidth]{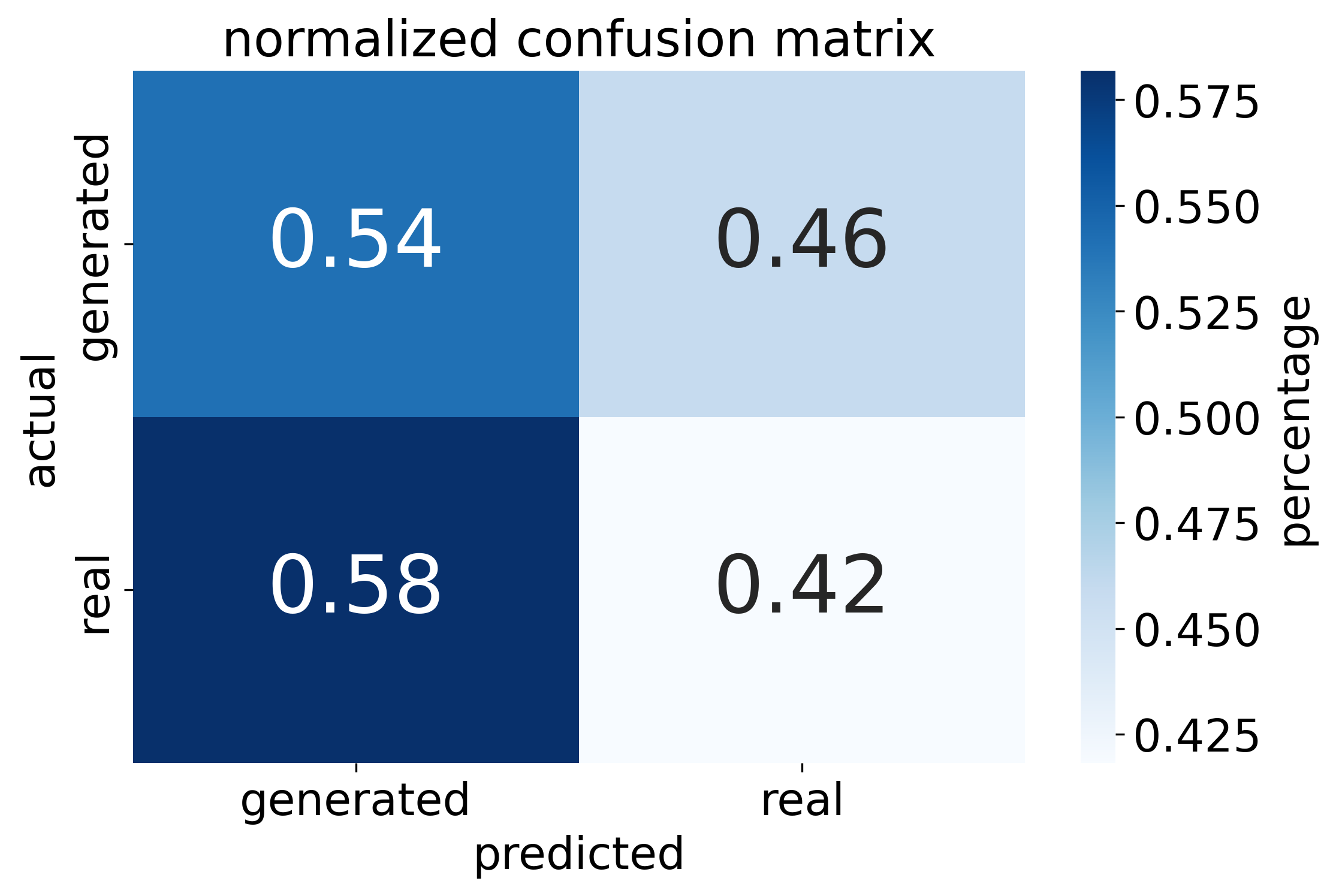}
    \caption{Normalized confusion matrix of expert survey classifications. Overall accuracy is approximately 50\%, demonstrating the significant difficulty for experts in distinguishing real and synthetic brightfield microscopy images.}
    \label{fig:conf_matrix}
\end{figure}

\begin{figure}
    \centering
    \includegraphics[width=0.2\linewidth]{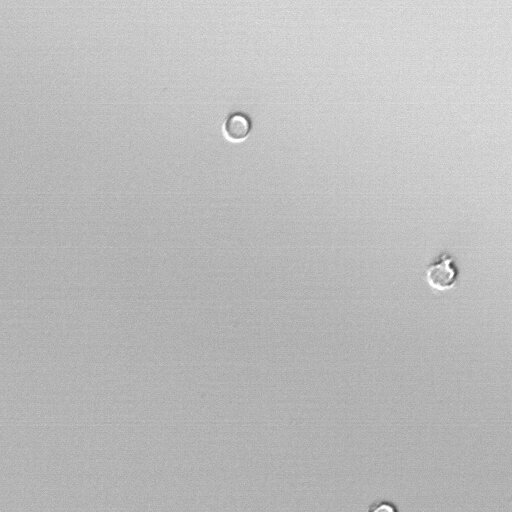}
    \quad
    \includegraphics[width=0.2\linewidth]{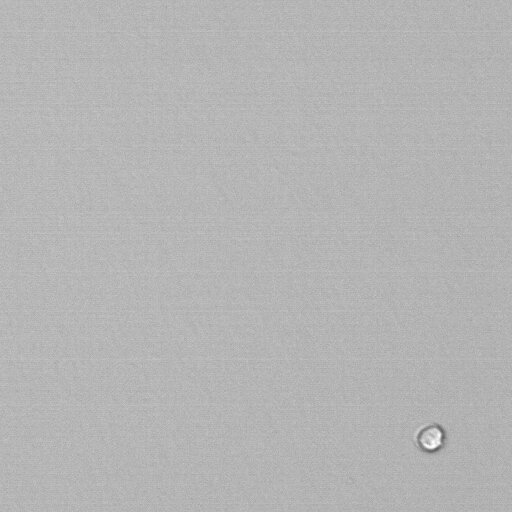}
    \qquad
    \qquad
    \includegraphics[width=0.2\linewidth]{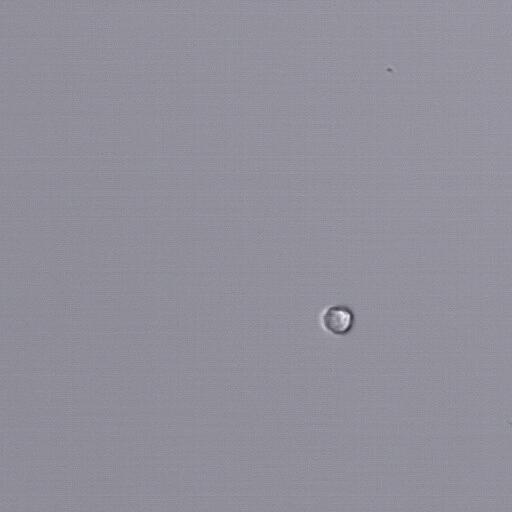}
    \quad
    \includegraphics[width=0.2\linewidth]{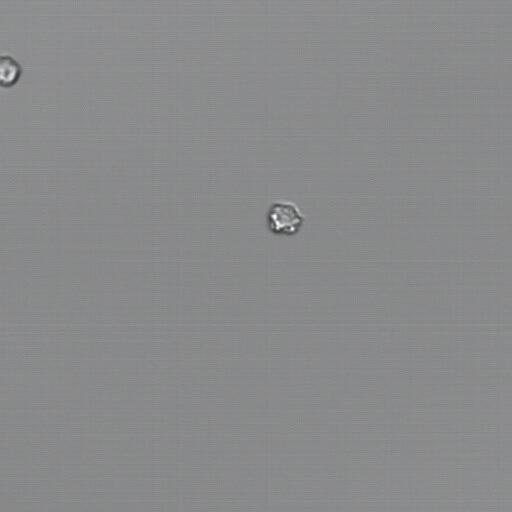}
    \caption{Visual comparison between real-world samples (left) and generated samples (right)}
    \label{fig:samples}
\end{figure}

The expert survey results revealed a striking inability of microscopy experts to consistently distinguish synthetic brightfield microscopy images generated by our diffusion model from real images.
The overall classification accuracy of all 11 participants and 30 images was remarkably close to chance level, reaching only 50\%.
This near-random performance strongly indicates the high degree of realism achieved by our synthetic image generation approach.
Figure~\ref{fig:conf_matrix} presents the normalized confusion matrix, visually demonstrating the balanced distribution of correct and incorrect classifications for both real and generated images.
Individual image accuracies varied significantly, ranging from approximately 18\% to 90\%, highlighting the inherent variability within both real and synthetic image sets and the complexity of the discrimination task.
Analysis of expert explanations revealed that experts considered subtle image features related to cell appearance, background texture, edge clarity, and perceived noise or artifacts when attempting to differentiate the images.
Prominent terms like ``cell,'' ``background,'' and ``edge'' indicate a focus on core microscopy image components, while terms such as ``pattern,'' ``perfect,'' ``artifacts,'' ``blurry,'' and ``contrast'' suggested that experts were searching for subtle imperfections or stylistic inconsistencies that might betray a synthetic origin.
However, these cues proved unreliable, ultimately leading to the near-chance level classification accuracy and underscoring the perceptual realism of our diffusion-generated brightfield microscopy images (Figure~\ref{fig:samples}).

\subsection{Object Detection Performance}
\label{subsec:object-detection-performance}
Tables~\ref{tab:model-performances1} and \ref{tab:model-performances2} summarize the object detection performance, presenting mAP metrics for all trained models across the seven datasets.
Analyzing mAP\@50, we observe that models trained with synthetic data replacing parts of the real data (\textit{scc\_10}, \textit{scc\_30}, \textit{scc\_50}) achieve comparable, and in some cases, improved performance compared to models trained solely on real data (\textit{scc\_real}).
Specifically, the YOLOv8s model trained on the \textit{scc\_30} dataset achieved a mAP\@50 of 0.9043, exceeding the 0.8947 achieved by the same model trained on the \textit{scc\_real} dataset.
Similarly, the RT-DETR-l model trained on the \textit{scc\_10} dataset reached a mAP\@50 of 0.9147, outperforming the 0.9146 achieved on the \textit{scc\_real} dataset.
However, when examining performance at higher IoU thresholds, specifically mAP\@75 and mAP\@50:95, a subtle trend of performance decrease emerges for models trained with increasing proportions of replaced synthetic data.
This trend is more pronounced for the RT-DETR models, indicating a potentially greater sensitivity of transformer-based architectures to the characteristics of synthetic data.

Continuing the analysis of object detection performance, we turn on the datasets where synthetic images were \textbf{added} to the real training data (\textit{scc\_add\_10}, \textit{scc\_add\_30}, \textit{scc\_add\_50}).
Examining mAP\@50 in Tables~\ref{tab:model-performances1} and \ref{tab:model-performances2}, we observe that models trained on these augmented datasets generally maintain a high level of performance, comparable to the baseline performance achieved with the \textit{scc\_real} dataset as well.
For instance, the YOLOv8s model shows a consistent mAP@50 around 0.8949-0.8952 across \textit{scc\_add\_10}, \textit{scc\_add\_30}, and \textit{scc\_add\_50}, similar to the 0.8947 of \textit{scc\_real}.
Notably, the YOLOv8m model achieves a mAP@50 of 0.9034 on \textit{scc\_add\_30}, closely matching the 0.9038 of the \textit{scc\_real} baseline.
The larger models, YOLOv8x, YOLOv9c, and YOLOv9e, also exhibit this trend of maintained high mAP@50 when trained with added synthetic data, suggesting that supplementing real data with synthetic images does not degrade the detection accuracy at this IoU threshold.

When considering the more stringent mAP@75 metric, we see a different pattern of maintained performance with added synthetic data compared to the replacement datasets.
In several instances, models trained on \textit{scc\_add\_30} even achieve slightly improved mAP@75 scores compared to the \textit{scc\_real} dataset.
For example, YOLOv8s reaches 0.8115 on \textit{scc\_add\_30} versus 0.8095 on \textit{scc\_real}, and YOLOv8m achieves 0.8224 on \textit{scc\_add\_30} compared to 0.8203 on \textit{scc\_real}.
This marginal improvement is also seen in YOLOv8x and YOLOv9c models on \textit{scc\_add\_30}.
Looking at the mAP@50:95 metric, only the YOLOv9e model trained on \textit{scc\_add\_50} achieves a higher score than the \textit{scc\_real} baseline, with a value of 0.6651 compared to 0.6629.

Generally, all models that were trained with replaced synthetic data augmentation maintained or achieved similar performance (less than 2\% difference) throughout the different datasets and IoU thresholds.
When it comes to models trained with added synthetic data, the performance was also maintained or slightly improved, particularly at lower IoU thresholds, with the exception of the RT-DETR models that showed a more pronounced sensitivity to the synthetic data proportion, achieving a
decreased performance of up to 5\% with the \textit{scc\_add\_50} dataset at more stringent IoU thresholds.
These results suggest that synthetic data augmentation can mostly maintain or even enhance detection performance at lower IoU thresholds, potentially due to increased data variability, dataset size and improved model robustness.

Qualitative assessment of sample detections revealed that all models, including those trained with synthetic data, generally performed well in detecting cells under the conditions proposed by the test split, including overlapping cells and cells at image boundaries.
However, subtle differences in detection precision and confidence scores were observed, especially for the transformer-based models, warranting further investigation into the fine-grained impact of synthetic data on model behavior.
\begin{table}[h!]
    \centering
    \caption{Detailed comparison of the performance metrics for the evaluated YOLO models across the different training datasets.}
    \label{tab:model-performances1}
    \resizebox{\linewidth}{!}{%
{\tiny
    \begin{tabular}{l|l|c|c|c}
        \toprule
        Model                         & Dataset                 & mAP @50                    & mAP @75                    & mAP @50:95                 \\
        \midrule
        \multirow[t]{7}{*}{YOLOv8s}   & \textit{scc\_real}      & 0.8947                     & 0.8095                     & \cellcolor{blue!10}0.6557  \\
                                      & \textit{scc\_10}        & 0.8941                     & 0.8053                     & 0.6470                     \\
                                      & \textit{scc\_30}        & \cellcolor{blue!10}0.9043  & 0.8079                     & 0.6448                     \\
                                      & \textit{scc\_50}        & 0.8932                     & 0.7891                     & 0.6325                     \\
                                      & \textit{scc\_add\_10}   & 0.8949                     & 0.8068                     & 0.6502                     \\
                                      & \textit{scc\_add\_30}   & 0.8952                     & \cellcolor{blue!10}0.8115  & 0.6536                     \\
                                      & \textit{scc\_add\_50}   & 0.8940                     & 0.8100                     & 0.6514                     \\
        \midrule
        \multirow[t]{7}{*}{YOLOv8m}   & \textit{scc\_real}      & \cellcolor{blue!10}0.9038  & 0.8203                     & \cellcolor{blue!10}0.6637  \\
                                      & \textit{scc\_10}        & 0.9035                     & 0.8093                     & 0.6558                     \\
                                      & \textit{scc\_30}        & 0.8945                     & 0.8076                     & 0.6462                     \\
                                      & \textit{scc\_50}        & 0.8930                     & 0.8040                     & 0.6456                     \\
                                      & \textit{scc\_add\_10}   & 0.8943                     & 0.8088                     & 0.6550                     \\
                                      & \textit{scc\_add\_30}   & 0.9034                     & \cellcolor{blue!10}0.8224  & 0.6606                     \\
                                      & \textit{scc\_add\_50}   & 0.8950                     & 0.8218                     & 0.6588                     \\
        \midrule
        \multirow[t]{7}{*}{YOLOv8x}   & \textit{scc\_real}      & \cellcolor{blue!10}0.9038  & 0.8120                     & \cellcolor{blue!10}0.6639  \\
                                      & \textit{scc\_10}        & 0.9032                     & 0.8055                     & 0.6531                     \\
                                      & \textit{scc\_30}        & 0.9031                     & 0.8085                     & 0.6549                     \\
                                      & \textit{scc\_50}        & 0.8935                     & 0.7961                     & 0.6425                     \\
                                      & \textit{scc\_add\_10}   & 0.8941                     & 0.8090                     & 0.6560                     \\
                                      & \textit{scc\_add\_30}   & 0.8942                     & \cellcolor{blue!10}0.8212  & \cellcolor{blue!10}0.6639  \\
                                      & \textit{scc\_add\_50}   & 0.8938                     & 0.8195                     & 0.6440                     \\
        \midrule
        \multirow[t]{7}{*}{YOLOv9c}   & \textit{scc\_real}      & \cellcolor{blue!10}0.9047  & 0.8215                     & \cellcolor{blue!10}0.6645  \\
                                      & \textit{scc\_10}        & 0.9042                     & 0.8070                     & 0.6531                     \\
                                      & \textit{scc\_30}        & 0.9042                     & 0.8106                     & 0.6568                     \\
                                      & \textit{scc\_50}        & 0.8935                     & 0.7951                     & 0.6422                     \\
                                      & \textit{scc\_add\_10}   & 0.8954                     & 0.8211                     & 0.6615                     \\
                                      & \textit{scc\_add\_30}   & 0.8952                     & \cellcolor{blue!10}0.8217  & 0.6605                     \\
                                      & \textit{scc\_add\_50}   & 0.8947                     & 0.8102                     & 0.6594                     \\
        \midrule
        \multirow[t]{7}{*}{YOLOv9e}   & \textit{scc\_real}      & 0.9037                     & 0.8201                     & 0.6629                     \\
                                      & \textit{scc\_10}        & \cellcolor{blue!10}0.9041  & 0.8141                     & 0.6650                     \\
                                      & \textit{scc\_30}        & 0.8946                     & 0.8069                     & 0.6485                     \\
                                      & \textit{scc\_50}        & 0.8938                     & 0.8037                     & 0.6453                     \\
                                      & \textit{scc\_add\_10}   & 0.8950                     & 0.8188                     & 0.6623                     \\
                                      & \textit{scc\_add\_30}   & 0.8952                     & \cellcolor{blue!10}0.8209  & 0.6638                     \\
                                      & \textit{scc\_add\_50}   & 0.9031                     & 0.8188                     & \cellcolor{blue!10}0.6651  \\
        \bottomrule
    \end{tabular}%
}
    }
\end{table}

\begin{table}[h!]
    \centering
    \caption{Detailed comparison of the performance metrics for the evaluated RT-DETR models across the different training datasets.}
    \label{tab:model-performances2}
    \resizebox{\linewidth}{!}{%
{\tiny
    \begin{tabular}{l|l|c|c|c}
        \toprule
        Model                         & Dataset                 & mAP @50                    & mAP @75                    & mAP @50:95                 \\
        \midrule
        \multirow[t]{7}{*}{RT-DETR-l} & \textit{scc\_real}      & 0.9146                     & \cellcolor{blue!10}0.8169  & \cellcolor{blue!10}0.6614  \\
                                      & \textit{scc\_10}        & \cellcolor{blue!10}0.9147  & 0.8071                     & 0.6574                     \\
                                      & \textit{scc\_30}        & 0.9017                     & 0.7780                     & 0.6240                     \\
                                      & \textit{scc\_50}        & 0.9036                     & 0.7843                     & 0.6298                     \\
                                      & \textit{scc\_add\_10}   & 0.9146                     & 0.8064                     & 0.6526                     \\
                                      & \textit{scc\_add\_30}   & 0.9043                     & 0.7962                     & 0.6457                     \\
                                      & \textit{scc\_add\_50}   & 0.9036                     & 0.7932                     & 0.6372                     \\
        \midrule
        \multirow[t]{7}{*}{RT-DETR-x} & \textit{scc\_real}      & \cellcolor{blue!10}0.9164  & \cellcolor{blue!10}0.8257  & \cellcolor{blue!10}0.6748  \\
                                      & \textit{scc\_10}        & 0.9144                     & 0.8083                     & 0.6565                     \\
                                      & \textit{scc\_30}        & 0.9032                     & 0.7830                     & 0.6328                     \\
                                      & \textit{scc\_50}        & 0.9045                     & 0.8032                     & 0.6437                     \\
                                      & \textit{scc\_add\_10}   & 0.9133                     & 0.7991                     & 0.6459                     \\
                                      & \textit{scc\_add\_30}   & 0.9136                     & 0.8106                     & 0.6532                     \\
                                      & \textit{scc\_add\_50}   & 0.9012                     & 0.7682                     & 0.6203                     \\
        \bottomrule
    \end{tabular}%
}
    }
\end{table}

\section{Discussion}
\label{sec:discussion}
The expert survey results provide compelling evidence for the high perceptual realism of brightfield microscopy images generated by our unconditional diffusion model.
The near-chance level accuracy achieved by microscopy experts in distinguishing synthetic images from real ones strongly supports the notion that diffusion models can synthesize microscopy data that is visually indistinguishable from real-world acquisitions.
This addresses our first sub-research question and highlights the potential of diffusion models to generate data suitable for augmenting or even substituting real microscopy images in certain applications.

Object detection experiments further reveal the practical utility of diffusion-based synthetic data for single-cell detection and, importantly, highlight the benefits of increasing dataset size.
The comparable, and in some cases, improved performance of models trained with synthetic data—both when replacing parts of the real data (\textbf{replacement datasets})
and when adding to it (\textbf{augmentation datasets})—demonstrates at mAP\@50 that synthetic images can effectively capture essential cell features.
This directly addresses our main research question, suggesting that synthetic data is a valuable asset in training robust cell detection models, particularly when labeled real data is limited.

Specifically, the results from the \textbf{augmentation datasets} underscore the advantage of expanding a dataset with synthetic data.
We observed that models trained on these augmented datasets consistently achieved performance at least comparable to, and often slightly better than, the baseline across various metrics, especially mAP\@50 and mAP\@75.
This improvement, while sometimes marginal, is notable considering it comes without additional real image acquisition.
The larger, more diverse training set likely allowed the models to learn more robust features and generalize better to unseen test data, as the added synthetic data effectively contributed to learning without introducing detrimental artifacts.

The subtle performance enhancements might be attributed to the increased variability from the synthetic data generation process, which improves model generalization to variations in real-world images.
Furthermore, the larger dataset size inherently provides more training examples, potentially leading to better parameter optimization.
This is consistent with the principle that larger, relevant datasets often lead to improved deep learning model performance—a point reinforced by the expert survey confirming the realism of our synthetic images.

However, a limitation emerges at higher IoU thresholds (mAP\@75, mAP\@50:95).
The subtle performance decrease observed with higher proportions of synthetic data, even in the \textbf{augmentation datasets}, indicates a potential deficit in the fidelity of synthetic images for capturing fine cell boundary details.
While adding synthetic data generally benefits performance by increasing dataset size, its inherent limitations in replicating all nuances of real microscopy data remain a factor for tasks requiring high localization precision.
The observed sensitivity of RT-DETR models to synthetic data proportions also warrants further investigation, potentially pointing to architectural differences in how transformer-based models leverage synthetic data compared to CNN-based models.

The implications for biological research are significant.
The ability to generate high-quality synthetic images and use them to train cell detection models opens new avenues for addressing data scarcity and annotation bottlenecks.
This offers a cost-effective and scalable approach to augment limited datasets, potentially democratizing access to advanced cell detection techniques.
Synthetic data can reduce dependence on time-consuming manual labor and, by enabling robust label-free detection, aligns with best practices for minimizing phototoxicity in live-cell imaging.
Moreover, it allows for the creation of datasets with rare cell phenotypes or challenging conditions that are difficult to capture experimentally.
For these benefits to be realized responsibly, transparency in reporting the use of synthetic data is crucial for maintaining scientific integrity.

\section{Conclusion}
\label{sec:conclusion}
This paper provides a comprehensive investigation into the use of diffusion-based synthetic brightfield microscopy images for enhancing single cell detection.
Our expert survey demonstrates the remarkable realism of diffusion-generated images, achieving near-indistinguishability from real microscopy acquisitions.
Object detection experiments reveal that models trained with synthetic data, especially when added to real data in \textbf{augmentation datasets} to increase dataset size, achieve comparable, and in some cases,
improved performance to real-data training, particularly for simpler cell localization (mAP\@50) and even showing benefits for higher IoU thresholds in some augmentation scenarios.
While subtle limitations exist in  replicating fine cell boundary details and achieving optimal performance at the highest IoU thresholds, our findings strongly highlight the promise of diffusion-based synthetic data generation as a valuable tool for microscopy image analysis.
It is important to note that this study was primarily designed to evaluate the general potential of synthetic data for training single-cell detection models, rather than to achieve state-of-the-art performance.
Consequently, we did not perform extensive hyperparameter tuning or explore advanced data augmentation techniques beyond the default settings provided by the Ultralytics framework.
This approach offers a pathway to address data scarcity, reduce annotation burdens, and potentially improve the robustness and accessibility of advanced cell detection techniques in biological and medical research,
particularly by leveraging the benefits of increased dataset size through synthetic data augmentation.
Future research directions should focus on refining diffusion models for brightfield microscopy image generation, improving the fidelity in capturing fine cellular details,
exploring conditional generation strategies for broader applicability across diverse microscopy modalities and biological contexts, and further investigating the optimal strategies for integrating synthetic data into training pipelines
to maximize the benefits for cell detection and other microscopy image analysis tasks, including exploring the optimal balance between real and synthetic data and the impact of dataset size scaling with synthetic data,
as well as investigating the impact of more extensive hyperparameter optimization and advanced data augmentation strategies when using synthetic data.

The source code is available on GitHub:
\url{https://github.com/mario-dg/Diffusion-Based-Brightfield-Image-Generation}.
The datasets and trained models are available on Hugging Face at \url{https://huggingface.co/mario-dg}.

{
\bibliographystyle{plain}
 \bibliography{bibliography}
}
\end{document}